\pdfoutput=1
\documentclass{article}

\usepackage[final,main]{neurips_2026}

\usepackage[utf8]{inputenc}
\usepackage[T1]{fontenc}
\usepackage[hidelinks]{hyperref}
\hypersetup{pdftitle={What Does an Observability Foundation Model Know?},pdfauthor={Dhyey Dharmendrakumar Mavani, Rian Atri, Tairan Ji}}
\usepackage{url}
\usepackage{booktabs}
\usepackage{amsfonts}
\usepackage{amsmath}
\usepackage{amssymb}
\usepackage{nicefrac}
\usepackage{microtype}
\usepackage{xcolor}
\usepackage{graphicx}
\usepackage{tabularx}
\usepackage{placeins}
\usepackage{flafter}
\usepackage{needspace}

\usepackage{array}
\usepackage{enumitem}
\usepackage{multirow}
\usepackage{tikz}
\usetikzlibrary{arrows.meta,calc,positioning,shapes.geometric,shapes.symbols,shapes.misc,fit,backgrounds,decorations.pathreplacing,patterns,patterns.meta}
\usepackage{pgfplots,pgfplotstable}
\usepgfplotslibrary{groupplots}
\ifdefined\crstyleloaded \fi\def\crstyleloaded{}
\usepgfplotslibrary{groupplots}
\usetikzlibrary{patterns, patterns.meta, decorations.pathreplacing}
\pgfplotsset{compat=1.18}

\definecolor{crToto}{HTML}{1D4E7A}
\definecolor{crTotoLight}{HTML}{3F88C5}
\definecolor{crMoment}{HTML}{7E2F24}
\definecolor{crWarm}{HTML}{C96F3B}
\definecolor{crGrayDark}{HTML}{A3A3A3}
\definecolor{crGrayLight}{HTML}{E2E2E2}
\definecolor{crInk}{HTML}{1A1A1A}
\definecolor{crMuted}{HTML}{595959}
\definecolor{crRule}{HTML}{BDBDBD}
\definecolor{crGrid}{HTML}{E6E6E6}
\definecolor{crPanel}{HTML}{F4F4F2}

\tikzset{
  crev5/.style={fill=crToto, text=white},
  crev4/.style={fill=crTotoLight, text=white},
  crev3/.style={fill=crGrayDark, text=crInk},
  crev2/.style={fill=crGrayLight, text=crInk},
  crev1/.style={fill=crWarm, text=crInk},
  crev0/.style={fill=crMoment, text=white},
  crevcell/.style={rounded corners=1.5pt, font=\footnotesize,
    align=center, inner sep=2pt},
  cresmall/.style={font=\scriptsize, text=crMuted},
}

\pgfplotsset{
  cr axis/.style={
    axis lines*=left,
    tick label style={font=\footnotesize},
    label style={font=\footnotesize},
    legend style={font=\scriptsize, draw=none, fill=none},
    grid style={line width=0.3pt, color=crGrid},
    major tick length=2pt,
  },
}

\usepackage{caption}
\usepackage{subcaption}
\usepackage{pifont}

\makeatletter
\renewcommand\paragraph{\@startsection{paragraph}{4}{\z@}{1.6ex \@plus 0.5ex \@minus 0.2ex}{0.4ex}{\normalfont\normalsize\bfseries}}
\makeatother

\title{What Does an Observability Foundation Model Know?}

\author{%
  Dhyey Dharmendrakumar Mavani\\
  Independent\\
  \texttt{ddmavani2003@gmail.com}
  \And
  Rian Atri\\
  Keiji AI\\
  \texttt{rianatri@keiji.ai}
  \And
  Tairan Ji\\
  Independent\\
  \texttt{tairanjiryan@gmail.com}
}

\begin{document}

\maketitle

\begin{abstract}
A linear probe can show that a label is recoverable from a model's hidden states, but not whether that goes beyond what the input already reveals, or whether the model uses it. We audit Toto, an observability forecasting foundation model, on the Benchmark of Observability Metrics (BOOM) across five series-disjoint resplits, comparing linear probes on its frozen residual stream with models that read the raw input window and with Toto's architecture stripped of its trained configuration. Short-vs-medium cadence and metric type are more linearly recoverable from Toto's residuals than from the strongest raw-window model in every resplit (macro-$F_1$ $0.766$ vs.\ $0.633$ and $0.545$ vs.\ $0.498$). Domain is nearly tied, and series cardinality is recovered far better from the raw window. MOMENT-base shows related cadence, metric-type, and domain readouts. Recoverability is not use: exchanging Toto's residuals with those of high-burst donors moves a future-burstiness readout as intended but does not make forecasts consistently burstier than a randomized donor. A BOOM-trained coordination probe has negative zero-shot $R^2$ on the tested external benchmarks. We report each label against its strongest baseline.
\end{abstract}

\section{Introduction}

Time-series foundation models (TSFMs) serve increasingly as zero-shot forecasters, and probing their hidden states is a natural way to ask what they have learned. Consider two telemetry series: request latency sampled every few seconds and disk usage sampled every few minutes, and suppose a linear probe on a forecaster's hidden states tells them apart. The model may have learned sampling cadence in pretraining. Or the two series may simply look different in the raw window, for instance in their step-to-step variation, so that any classifier on the input would separate them just as well. And even if the hidden states separate them better than the input does, the forecaster need not use that distinction when it predicts the next values. A probe score alone cannot tell these cases apart.

\begin{figure}[t]
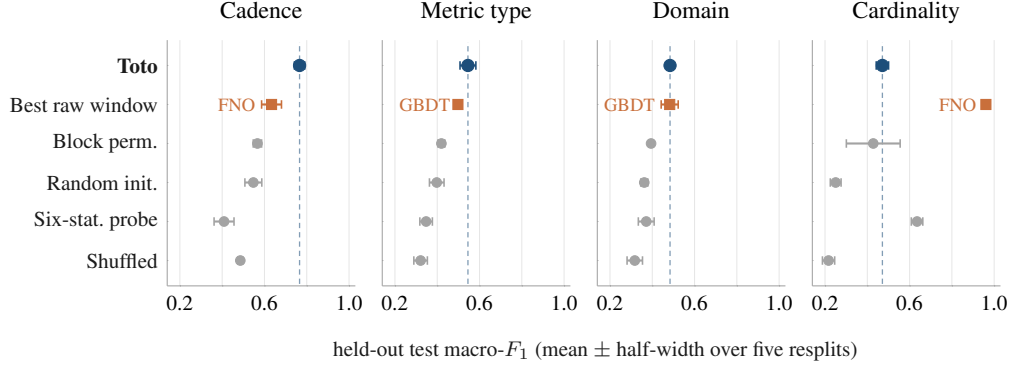

\centering
\input{figures/tikz/cr_style.tex}
\begin{tikzpicture}
\pgfplotsset{
  fig1 panel/.style={cr axis,
    scale only axis, width=0.98in, height=1.28in,
    xmin=0.14, xmax=1.03, ymin=0.35, ymax=6.65,
    xtick={0.2,0.4,0.6,0.8,1.0}, xticklabels={0.2,,0.6,,1.0},
    xmajorgrids=true, ytick={1,...,6}, yticklabels=\empty,
    tick label style={font=\footnotesize},
    major tick length=0pt, axis line style={crGrayDark, line width=0.4pt},
    title style={font=\small, text height=1.6ex, text depth=0.4ex, yshift=-3pt}, clip=false,
  },
  fig1 errs/.style={error bars/.cd, x dir=both, x explicit,
    error bar style={line width=0.8pt}, error mark=|, error mark options={mark size=1.6pt, line width=0.6pt}},
  fig1 toto/.style={only marks, mark=*, mark size=2.3pt, color=crToto, fig1 errs},
  fig1 raw/.style={only marks, mark=square*, mark size=1.9pt, color=crWarm, fig1 errs},
  fig1 control/.style={only marks, mark=*, mark size=1.8pt, color=crGrayDark, fig1 errs},
}
\tikzset{
  fig1 ref/.style={crToto, line width=0.5pt, dash pattern=on 2pt off 1.5pt, opacity=0.55},
  fig1 fam/.style={font=\scriptsize, text=crWarm, inner sep=1pt},
}
\begin{axis}[fig1 panel, at={(0.98in,0in)}, anchor=south west, title={Cadence},
  yticklabels={Shuffled, Six-stat.\ probe, Random init., Block perm., Best raw window, \textbf{Toto}},
  yticklabel style={font=\footnotesize, text=crInk}]
\input{figures/tikz/gen/fig1_panel1.tex}
\end{axis}
\begin{axis}[fig1 panel, at={(2.10in,0in)}, anchor=south west, title={Metric type}]
\input{figures/tikz/gen/fig1_panel2.tex}
\end{axis}
\begin{axis}[fig1 panel, at={(3.22in,0in)}, anchor=south west, title={Domain}]
\input{figures/tikz/gen/fig1_panel3.tex}
\end{axis}
\begin{axis}[fig1 panel, at={(4.34in,0in)}, anchor=south west, title={Cardinality}]
\input{figures/tikz/gen/fig1_panel4.tex}
\end{axis}
\node[font=\footnotesize, text=crInk, anchor=north] at (3.21in,-0.24in) {held-out test macro-$F_1$ (mean $\pm$ half-width over five resplits)};
\end{tikzpicture}
\caption{Held-out test macro-$F_1$ of Toto, its baselines, and the shuffled-label floor for each taxonomy label (means $\pm$ half-widths over five resplits; \S\ref{sec:method}). The dashed line marks Toto's mean; the orange square is the strongest raw-window model for that label.}
\label{fig:evidence-map}
\end{figure}

We study this question for Toto \citep{datadog2024toto,cohen2025observability}, an observability forecasting foundation model, on the Benchmark of Observability Metrics (BOOM) \citep{datadog2025boom}, following probing work that separates decodability from use \citep{tenney2019bert,tenney2019probing,conneau2018sentence,rogers2020primer,belinkov2022probing,hewitt2019designing,meng2022locating,turner2024steering,geiger2023finding,geiger2025theoretical,olsson2022incontext}. We read ``know'' operationally: a label is known to the extent that a linear probe on Toto's frozen residual stream predicts it on held-out series. This is a claim about linear recoverability from one representation under one probe family, not about semantic understanding or all the information the model contains. For each label we ask three questions. Is it more linearly recoverable from Toto's residuals than from the input window itself? Does that depend on Toto's trained weights and their order? And does moving the readout move the forecast? \emph{Input baselines}---a probe on six summary statistics of the input and four models trained on the raw window---answer the first. \emph{Backbone baselines}, Toto with random weights or with its trained blocks in shuffled order, answer the second. A donor exchange tests the third on future burstiness, and zero-shot probe transfer asks whether a probe fitted on BOOM carries over unchanged to external datasets. Figure~\ref{fig:evidence-map} summarizes the first two questions for each label.

\paragraph{Contributions}
First, we give a simple audit design for forecasting representations: every held-out probe score is paired, on the same windows, with input and backbone baselines, and a separate donor exchange tests forecast response. Second, we report a label-by-label result on BOOM across five series-disjoint resplits: cadence and metric type are more linearly recoverable from Toto's residuals than from every tested raw-window model in all five resplits, and their gaps over the six-statistic probe and random initialization persist within the tested strata of the other labels; domain is nearly tied with its strongest raw-window model, and cardinality is recovered far better from the raw window. Third, we bound how far probe readouts can be interpreted. MOMENT-base shows related cadence, metric-type, and domain readouts on the same resplits, with narrower baselines. In both models, exchanging residuals moves the targeted future-burstiness readout in every resplit without consistently moving the forecast endpoint---forecast burstiness for Toto and future mean absolute error (MAE) for MOMENT---and a BOOM-trained coordination probe has negative zero-shot $R^2$ on the tested external benchmarks. We never treat a moved readout as evidence that the model uses it to forecast.

\section{Related Work}
\label{sec:related}

\paragraph{Forecasting foundation models and observability}
TSFMs differ in pretraining objective and use: Chronos and TimesFM cast broad-domain forecasting as autoregressive prediction, whereas MOMENT is a general-purpose encoder trained with masked reconstruction \citep{ansari2024chronos,das2024timesfm,goswami2024moment}. Toto adapts the forecasting-foundation-model paradigm to multivariate telemetry, with the accompanying BOOM benchmark supplying real observability series and forecast tasks \citep{datadog2024toto,cohen2025observability,datadog2025boom}. An interpretability audit asks instead whether frozen hidden states make series attributes accessible to a simple readout.

\paragraph{Probing, selectivity, and controls}
Linear probes have long asked which properties are recoverable from intermediate representations \citep{alain2016understanding,conneau2018sentence,tenney2019bert}; probe capacity and control tasks matter because a flexible probe can expose structure available without the representation under study \citep{hewitt2019designing,belinkov2022probing}. Recent TSFM work makes this question concrete: Wili\'nski et al. analyze layer structure and probe concepts such as trend and periodicity, then steer models with concept-derived interventions; Pandey et al. probe synthetic temporal factors and their compositions \citep{wilinski2025representations,pandey2025internalsemantics}. We ask the same question for telemetry taxonomy labels on real BOOM series, with raw-window neural and tree models over the same context as comparators.

\paragraph{Decodability, interventions, and model use}
A model need not use a readable direction to compute its output. Control-task probing makes this explicit, while causal tracing and editing, activation steering, and causal abstraction test how internal changes relate to behavior \citep{hewitt2019designing,meng2022locating,turner2024steering,geiger2023finding,geiger2025theoretical}. These methods come largely from language-model settings; we adopt their reading-vs-intervening distinction. Related TSFM work has already used concept-based interventions to alter temporal patterns and forecasts \citep{wilinski2025representations}. We use a donor exchange to test whether moving a readout moves a forecast.

\paragraph{Forecast evaluation and transfer}
Forecast Evaluation (FEV) emphasizes realistic task coverage, covariates, reproducible comparisons, and aggregation across tasks \citep{shchur2025fev}. Public long-horizon suites likewise enable cross-dataset forecast evaluation \citep{zhou2021informer,thuml2024tslib}. Our external test is narrower than either forecast transfer or representation transfer: it applies a BOOM-trained probe, unchanged, to external FEV configurations and long-sequence datasets.

\section{Setup}
\label{sec:setup}

\paragraph{Models}
Toto is a decoder-only transformer over patchified multivariate inputs \citep{datadog2024toto,nie2023patchtst,dosovitskiy2021vit}; we load the public \texttt{Datadog/Toto-Open-Base-1.0} checkpoint without fine-tuning. We divide each 1024-observation context into 16 non-overlapping 64-observation patches and read activations at the patch embedding and layers 0--11. The MOMENT-base encoder \citep{goswami2024moment} uses masked-reconstruction pretraining; we load it in embedding mode and read activations at layers 3, 6, 9, and 11. MOMENT uses a 512-observation context with 8-observation patches, so its six-statistic probe pools a different final patch than Toto's in Table~\ref{tab:structural-controls}. Further protocol details are in App.~\ref{app:protocol}.

\paragraph{Data}
BOOM has $2{,}807$ observability series and approximately $350$ million observations \citep{cohen2025observability,datadog2025boom}. The fixed context-and-target length excludes $489$ series ($17.4\%$), including daily and weekly cadences, so the cadence label here is a Short-vs-Medium bucket. All results use five seeded, series-disjoint resplits (seeds 42--46): a 70/15/15 split capped at 500 training series gives 500/421/422 train/validation/test series per resplit, leaving the rest unused. Test sets differ across resplits but can overlap. Series too short for the fixed context contribute no windows; up to four windows are sampled per remaining series (App.~\ref{app:protocol}). Zero-shot probe transfer uses 11 FEV configurations \citep{shchur2025fev} and four long-sequence time-series forecasting (LSTF) datasets (ETTh1, ETTh2, weather, electricity) \citep{thuml2024tslib,zhou2021informer}; the FEV configurations may share underlying sources (IDs in App.~\ref{app:protocol}).

\begin{figure}[t]
\centering
\input{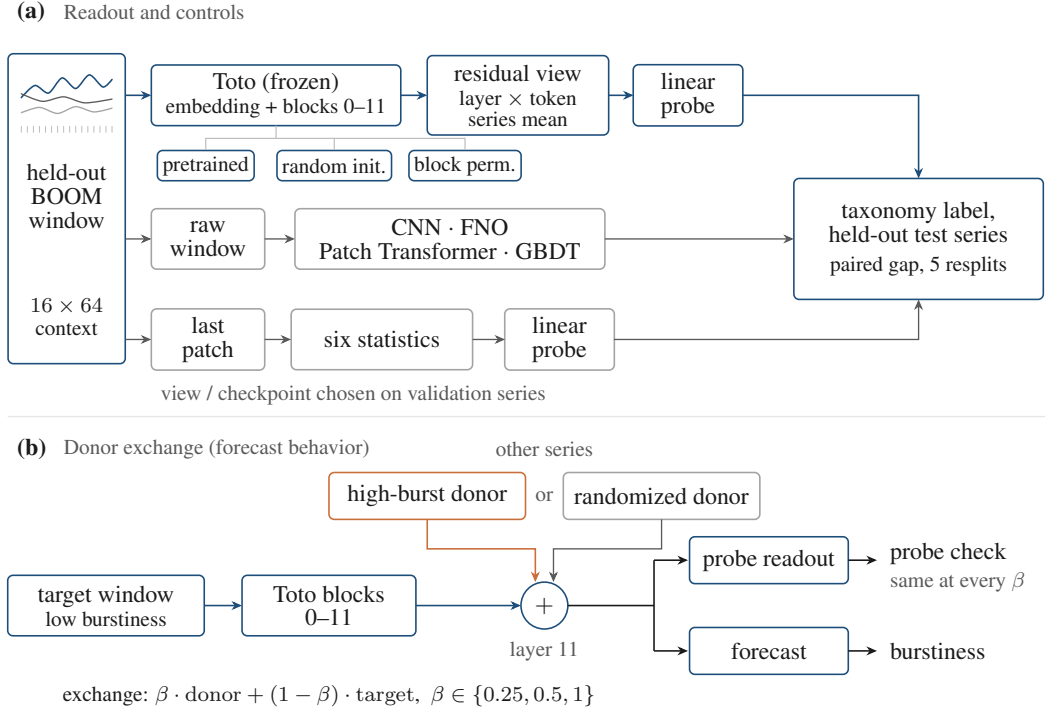}
\begin{tikzpicture}[x=1cm, y=1cm]
\tikzset{
  msbox/.style={draw, rounded corners=2pt, line width=0.6pt, inner sep=3pt,
    align=center, font=\small, text=crInk, fill=white},
  mstoto/.style={msbox, draw=crToto},
  msgray/.style={msbox, draw=crGrayDark},
  mswarm/.style={msbox, draw=crWarm},
  mssub/.style={font=\footnotesize, text=crInk},
  msmuted/.style={font=\footnotesize, text=crMuted, align=center},
  mstag/.style={font=\small\bfseries, text=crInk},
  msarr/.style={-{Stealth[length=4pt,width=3.4pt]}, line width=0.6pt, draw=crInk},
  mstarr/.style={-{Stealth[length=4pt,width=3.4pt]}, line width=0.6pt, draw=crToto},
  msgarr/.style={-{Stealth[length=4pt,width=3.4pt]}, line width=0.5pt, draw=crMuted},
  mswarr/.style={-{Stealth[length=4pt,width=3.4pt]}, line width=0.6pt, draw=crWarm},
}

\begin{scope}[yshift=4.35cm]
\node[mstag, anchor=west] at (0.0,5.05) {(a)};
\node[msmuted, anchor=west] at (0.62,5.05) {Readout and controls};

\node[mstoto, minimum width=1.55cm, minimum height=4.10cm] (input) at (0.78,2.45) {};
\begin{scope}[line join=round, line cap=round]
\draw[crToto, line width=0.6pt] plot[smooth] coordinates
  {(0.18,3.92)(0.36,4.08)(0.54,3.98)(0.72,4.18)(0.90,4.04)(1.08,4.22)(1.26,4.06)(1.38,4.16)};
\draw[crGrayDark, line width=0.5pt] plot[smooth] coordinates
  {(0.18,3.72)(0.42,3.79)(0.66,3.71)(0.90,3.80)(1.14,3.73)(1.38,3.78)};
\draw[crMuted, line width=0.5pt] plot[smooth] coordinates
  {(0.18,3.97)(0.46,3.87)(0.74,3.93)(1.02,3.85)(1.38,3.90)};
\foreach \i in {0,...,15}
  \draw[crRule, line width=0.35pt] ({0.18+\i*0.08},3.46) -- ({0.18+\i*0.08},3.54);
\end{scope}
\node[font=\small, text=crInk, align=center] at (0.78,2.62) {held-out\\[-1pt]BOOM\\[-1pt]window};
\node[mssub, align=center] at (0.78,1.02) {$16\times64$\\[-1pt]context};

\node[mstoto, minimum width=3.30cm, minimum height=0.80cm] (toto) at (3.55,3.95)
  {Toto (frozen)\\[-1pt]{\footnotesize embedding + blocks 0--11}};
\node[mstoto, minimum width=2.40cm, minimum height=1.05cm] (view) at (6.75,3.95)
  {residual view\\[-1pt]{\footnotesize layer $\times$ token}\\[-2pt]{\footnotesize series mean}};
\node[mstoto, minimum width=1.45cm, minimum height=0.80cm] (probe) at (9.00,3.95) {linear\\[-1pt]probe};
\draw[mstarr] (1.56,3.95) -- (1.90,3.95);
\draw[mstarr] (5.20,3.95) -- (5.55,3.95);
\draw[mstarr] (7.95,3.95) -- (8.28,3.95);
\node[msbox, draw=crToto, font=\footnotesize, inner sep=2pt, minimum height=0.40cm] (vpre) at (2.62,3.02) {pretrained};
\node[msbox, draw=crToto, font=\footnotesize, inner sep=2pt, minimum height=0.40cm] (vrand) at (4.32,3.02) {random init.};
\node[msbox, draw=crToto, font=\footnotesize, inner sep=2pt, minimum height=0.40cm] (vperm) at (6.05,3.02) {block perm.};
\draw[crRule, line width=0.45pt] (3.55,3.55) -- (3.55,3.38);
\draw[crRule, line width=0.45pt] (2.62,3.23) -- (2.62,3.38) -- (6.05,3.38) -- (6.05,3.23);
\draw[crRule, line width=0.45pt] (4.32,3.38) -- (4.32,3.23);

\node[msgray, minimum width=1.50cm, minimum height=0.80cm] (raw) at (2.65,2.05) {raw\\[-1pt]window};
\node[msgray, minimum width=4.10cm, minimum height=0.80cm] (models) at (5.85,2.05)
  {CNN $\cdot$ FNO\\[-1pt]Patch Transformer $\cdot$ GBDT};
\draw[msgarr] (1.56,2.05) -- (1.90,2.05);
\draw[msgarr] (3.40,2.05) -- (3.80,2.05);

\node[msgray, minimum width=1.50cm, minimum height=0.80cm] (last) at (2.65,0.72) {last\\[-1pt]patch};
\node[msgray, minimum width=2.40cm, minimum height=0.80cm] (six) at (4.95,0.72) {six statistics};
\node[msgray, minimum width=1.45cm, minimum height=0.80cm] (sprobe) at (7.30,0.72) {linear\\[-1pt]probe};
\draw[msgarr] (1.56,0.72) -- (1.90,0.72);
\draw[msgarr] (3.40,0.72) -- (3.75,0.72);
\draw[msgarr] (6.15,0.72) -- (6.58,0.72);

\node[mstoto, minimum width=3.30cm, minimum height=1.60cm] (gap) at (12.05,2.05)
  {taxonomy label,\\[-1pt]held-out test series\\[1pt]{\footnotesize paired gap, 5 resplits}};
\draw[mstarr] (9.72,3.95) -- (12.05,3.95) -- (12.05,2.85);
\draw[msgarr] (7.90,2.05) -- (10.40,2.05);
\draw[msgarr] (8.02,0.72) -- (12.05,0.72) -- (12.05,1.25);
\node[msmuted, anchor=west] at (1.90,-0.02) {view / checkpoint chosen on validation series};
\end{scope}

\draw[crGrid, line width=0.5pt] (0.0,4.05) -- (13.9,4.05);

\node[mstag, anchor=west] at (0.0,3.62) {(b)};
\node[msmuted, anchor=west] at (0.62,3.62) {Donor exchange (forecast behavior)};

\node[msmuted] at (7.10,3.62) {other series};
\node[mswarm, minimum width=2.60cm, minimum height=0.62cm] (hb) at (5.55,3.00) {high-burst donor};
\node[msgray, minimum width=2.60cm, minimum height=0.62cm] (rnd) at (8.65,3.00) {randomized donor};
\node[msmuted] at (7.10,3.00) {or};

\node[mstoto, minimum width=2.60cm, minimum height=0.80cm] (tgt) at (1.30,1.55)
  {target window\\[-1pt]{\footnotesize low burstiness}};
\node[mstoto, minimum width=2.30cm, minimum height=0.80cm] (blk) at (4.25,1.55)
  {Toto blocks\\[-1pt]0--11};
\node[mstoto, circle, minimum size=0.62cm, inner sep=0pt, font=\small] (mix) at (7.10,1.55) {$+$};
\draw[mstarr] (2.60,1.55) -- (3.10,1.55);
\draw[mstarr] (5.40,1.55) -- (6.79,1.55);
\draw[mswarr] (5.55,2.69) -- (5.55,2.30) -- (6.98,2.30) -- (6.98,1.86);
\draw[msgarr] (8.65,2.69) -- (8.65,2.30) -- (7.22,2.30) -- (7.22,1.86);
\node[msmuted, anchor=north] at (7.10,1.20) {layer 11};
\node[mssub, anchor=north] at (4.25,0.62) {exchange: $\beta\cdot\mathrm{donor}+(1-\beta)\cdot\mathrm{target}$, \ $\beta\in\{0.25,0.5,1\}$};

\draw[line width=0.6pt, draw=crInk] (7.41,1.55) -- (8.55,1.55);
\draw[line width=0.6pt, draw=crInk] (8.55,2.15) -- (8.55,0.95);
\draw[msarr] (8.55,2.15) -- (9.00,2.15);
\draw[msarr] (8.55,0.95) -- (9.00,0.95);
\node[mstoto, minimum width=2.10cm, minimum height=0.62cm] (pread) at (10.07,2.15) {probe readout};
\node[mstoto, minimum width=2.10cm, minimum height=0.62cm] (fcst) at (10.07,0.95) {forecast};
\draw[msarr] (11.12,2.15) -- (11.50,2.15);
\draw[msarr] (11.12,0.95) -- (11.50,0.95);
\node[font=\small, text=crInk, anchor=west] at (11.55,2.24) {probe check};
\node[msmuted, anchor=west] at (11.55,1.86) {same at every $\beta$};
\node[font=\small, text=crInk, anchor=west] at (11.55,0.95) {burstiness};
\end{tikzpicture}
\caption{Method. (a) Linear probes on frozen Toto residual views (pretrained backbone, or the randomly initialized and block-permuted backbone baselines) versus the input baselines (raw-window models and a six-statistic probe), compared as paired gaps on held-out series. (b) Layer-11 donor exchange, read by the probe and the forecast.}
\label{fig:method}
\end{figure}

\paragraph{Pretraining/evaluation overlap}
Toto was pretrained on an internal observability corpus; overlap with BOOM is unknown. The series-disjoint resplits prevent within-study train/test leakage.

\paragraph{Labels}
We audit four BOOM-metadata taxonomy labels: cadence (sampling-frequency metadata, restricted to the Short and Medium buckets that fit the fixed context-and-target length), metric type (gauge, rate, distribution), domain (Application Usage, Infrastructure, Database), and series cardinality (variate count in bins 1, 2--8, 9--32, above 32). Other and multilabel metric/domain entries are outside these singleton-label probes.

\needspace{6\baselineskip}
Dynamic labels are engineered statistics of the context and the next patch. For each variate $i$, the robust scale $s_i$ is $1.4826$ times the median absolute deviation (MAD) of its context, clamped away from zero. Burstiness is the largest absolute deviation from the context median, divided by $s_i$, over the final context patch (current) or the true next patch (future). Sparsity is the fraction of exact zeros in the same two patches. Shift risk averages $|\bar{x}^{\mathrm{next}}_i-\bar{x}^{\mathrm{last}}_i|/s_i$ over all variates of the series.

\needspace{3\baselineskip}
For coordination, let $d_i$ be the shift of variate $i$'s mean from the final context patch to the next patch. A variate is active if $|d_i| > 0.5\,s_i$; coordination is the larger of the fractions of active variates that move up and that move down, and zero if no variate is active.

\section{Method}
\label{sec:method}

\paragraph{Held-out readouts}
Figure~\ref{fig:method} sketches the design. We fit logistic-regression probes for categorical labels and ridge probes for continuous labels, and report macro-$F_1$ and $R^2$. The candidate views are 13 layers by 3 token positions with series-mean pooling (39 views). We select the view by validation accuracy (macro-$F_1$ for cardinality) and keep probe regularization fixed (App.~\ref{app:protocol}).

\begin{table}[t]
\centering
\small
\caption{Held-out test macro-$F_1$ on BOOM (means $\pm$ half-widths over five resplits). A: Toto against the input baselines. The strongest raw-window model is chosen by mean validation macro-$F_1$; the paired gap is Toto minus that model, and wins count the resplits in which it is positive. B: backbone baselines and the shuffled-label floor. All raw-window models are in App.~\ref{app:raw-controls}.}
\label{tab:structural-controls}
\begin{tabular*}{\linewidth}{@{\extracolsep{\fill}}lccccc}
\toprule
\multicolumn{6}{@{}l}{\textbf{A. Input baselines}} \\
\midrule
Label & Toto & Six-stat.\ probe & \shortstack{Strongest\\raw-window model} & Paired gap & Wins \\
\midrule
Cadence & $0.766\pm0.024$ & $0.409\pm0.047$ & \makebox[3.3em][l]{FNO}$0.633\pm0.047$ & $+0.133\pm0.063$ & 5/5 \\
Metric type & $0.545\pm0.037$ & $0.347\pm0.030$ & \makebox[3.3em][l]{GBDT}$0.498\pm0.007$ & $+0.047\pm0.036$ & 5/5 \\
Domain & $0.484\pm0.011$ & $0.371\pm0.037$ & \makebox[3.3em][l]{GBDT}$0.482\pm0.040$ & $+0.002\pm0.040$ & 2/5 \\
Cardinality & $0.471\pm0.029$ & $0.635\pm0.027$ & \makebox[3.3em][l]{FNO}$0.961\pm0.016$ & $-0.490\pm0.044$ & 0/5 \\
\addlinespace[4pt]
\multicolumn{6}{@{}l}{\textbf{B. Backbone baselines and shuffled-label floor}} \\
\midrule
Label & Random init. & Block perm. & Shuffled labels & & \\
\midrule
Cadence & $0.547\pm0.040$ & $0.567\pm0.020$ & $0.486\pm0.011$ & & \\
Metric type & $0.398\pm0.035$ & $0.420\pm0.016$ & $0.321\pm0.032$ & & \\
Domain & $0.362\pm0.017$ & $0.394\pm0.011$ & $0.317\pm0.037$ & & \\
Cardinality & $0.250\pm0.025$ & $0.428\pm0.127$ & $0.217\pm0.029$ & & \\
\bottomrule
\end{tabular*}
\end{table}

\begin{figure}[t]
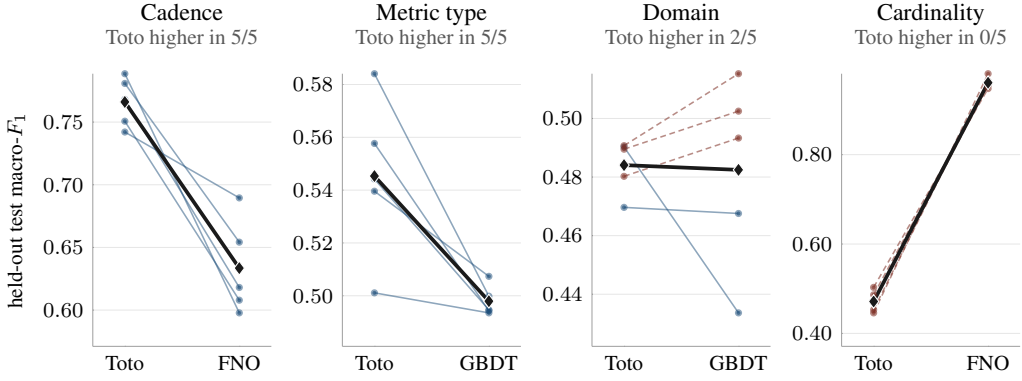

\centering
\input{figures/tikz/cr_style.tex}
\input{figures/tikz/gen/fig3_defs.tex}
\begin{tikzpicture}
\pgfplotsset{
  fig3 panel/.style={cr axis,
    scale only axis, width=0.93in, height=1.42in,
    xmin=-0.28, xmax=1.28, xtick={0,1},
    enlarge y limits={value=0.14, upper},
    enlarge y limits={value=0.14, lower},
    ymajorgrids=true, max space between ticks=26pt,
    yticklabel style={font=\footnotesize, /pgf/number format/fixed, /pgf/number format/fixed zerofill, /pgf/number format/precision=2},
    xticklabel style={font=\footnotesize},
    axis line style={crGrayDark, line width=0.4pt}, major tick length=0pt,
    title style={font=\small, align=center, text depth=0.4ex},
    clip=false,
  },
  fig3 up/.style={crToto, line width=0.6pt, opacity=0.5, mark=*, mark size=1.05pt,
    mark options={solid, fill opacity=0.8}},
  fig3 down/.style={crMoment, line width=0.6pt, opacity=0.55, dash pattern=on 2.2pt off 1.4pt,
    mark=*, mark size=1.05pt, mark options={solid, fill opacity=0.8}},
  fig3 mean/.style={crInk, line width=1.4pt, mark=diamond*, mark size=2.6pt,
    mark options={fill=crInk, draw=white, line width=0.4pt}},
}
\begin{axis}[fig3 panel, at={(0.54in,0in)}, anchor=south west,
  xticklabels={Toto,\figThreeFamA}, ylabel={held-out test macro-$F_1$}, ylabel style={font=\footnotesize},
  title={Cadence\\[-1pt]{\footnotesize\color{crMuted}Toto higher in \figThreeWinsA/5}}]
\input{figures/tikz/gen/fig3_panel1.tex}
\end{axis}
\begin{axis}[fig3 panel, at={(1.84in,0in)}, anchor=south west,
  xticklabels={Toto,\figThreeFamB},
  title={Metric type\\[-1pt]{\footnotesize\color{crMuted}Toto higher in \figThreeWinsB/5}}]
\input{figures/tikz/gen/fig3_panel2.tex}
\end{axis}
\begin{axis}[fig3 panel, at={(3.14in,0in)}, anchor=south west,
  xticklabels={Toto,\figThreeFamC},
  title={Domain\\[-1pt]{\footnotesize\color{crMuted}Toto higher in \figThreeWinsC/5}}]
\input{figures/tikz/gen/fig3_panel3.tex}
\end{axis}
\begin{axis}[fig3 panel, at={(4.44in,0in)}, anchor=south west,
  xticklabels={Toto,\figThreeFamD},
  title={Cardinality\\[-1pt]{\footnotesize\color{crMuted}Toto higher in \figThreeWinsD/5}}]
\input{figures/tikz/gen/fig3_panel4.tex}
\end{axis}
\end{tikzpicture}
\caption{Held-out test macro-$F_1$ in each resplit for Toto and the strongest raw-window model (thin lines: solid where Toto is higher, dashed where the raw-window model is higher) and their five-resplit means (bold). Each panel has its own vertical scale.}
\label{fig:resplit-slopes}
\end{figure}

\paragraph{Baselines}
\emph{Input baselines} ask whether a label is already recoverable from the input window. The six-statistic probe is a linear probe on the last 64-observation patch's mean, standard deviation, sparsity, mean absolute value, maximum absolute value, and cross-variate spread.

\needspace{4\baselineskip}
Four raw-window models are trained from scratch on the same windows: a convolutional network (CNN), a Fourier neural operator (FNO), and a patch Transformer, which receive the masked full context, with variates as input channels zero-padded to a common count, plus a coverage channel (the share of unpadded variates); and a gradient-boosted tree (GBDT) on 15 engineered features, including the six last-patch statistics and full-context spectral and autocorrelation summaries.

\needspace{4\baselineskip}
\emph{Backbone baselines} ask whether a readout depends on Toto's trained configuration: Toto with randomly initialized weights, and a block permutation that runs the 12 pretrained transformer blocks in a non-identity order (its view search also admits per-variate pooling, 78 views, which can only favor it). Both keep Toto's 64-observation patching, its parameter-free input scaling, and the same windows, so these shared components do not explain gaps over them. A shuffled-label probe on Toto's residuals gives a chance-level reference, the \emph{shuffled-label floor}.

The raw-window models are not evaluators of the same kind as the probe: they are nonlinear and trained from scratch on the 500 training series of each resplit, whereas the probe is linear on a representation pretrained elsewhere. A gap in Toto's favor therefore means that a label is more linearly recoverable from Toto's residuals than these models, with this data and training budget, learn to recover it from the window; the window itself may still carry the information.

For label $y$, baseline $c$, and resplit $s$, the paired macro-$F_1$ gap is $\Delta_{y,c,s}=F_{1}(y,\text{pretrained},s)-F_{1}(y,c,s)$. Each label's strongest raw-window model is the one with the highest mean validation macro-$F_1$ across the five resplits; the paired gap is computed against it in every resplit. Choosing by mean test macro-$F_1$ would select the same models, and for cadence and metric type the gap is positive against all four raw-window models in all five resplits.

\paragraph{Common-support probes}
Some taxonomy labels are correlated, so a readout of one could partly reflect another. Three prespecified common-support probes therefore retest the gaps over the six-statistic probe and random initialization within one fixed stratum of the other two labels. Each averages windows within series, balances target classes by series, selects the view on validation data, and measures test performance once. Raw-window models were not refit within strata.

\paragraph{Donor exchange}
Our donor exchange tests forecast behavior by blending residuals from real donor series into a target series' state. In each resplit we fit a future-burstiness probe on training series at the fixed layer-11, all-context, series-mean view, and evaluate 40 triples at residual blends $\beta\in\{0.25,0.5,1.0\}$. Each triple pairs a held-out target window from the lowest future-burstiness quartile with a high-burst donor (a top-quartile window from another series) and a randomized donor (any eligible test window from another series, not matched on covariates or taxonomy). Because the probe is linear, the difference between the two exchanges' probe scores equals $\beta$ times the difference between the two donors' probe scores. The \emph{probe check} is the fraction of targets for which the high-burst exchange raises the readout more than the randomized exchange. It is therefore the same at every blend. The check confirms that the exchange moves the readout in the intended direction; only the forecast endpoints vary with $\beta$. The primary forecast endpoint is whether the forecast is burstier under the high-burst donor than under the randomized donor; the secondary endpoint is the ratio of their weighted absolute percentage errors (WAPE).

\paragraph{Reporting conventions} The resplit is the unit of analysis: each resplit draws disjoint train, validation, and test series from the same BOOM corpus. Unless stated otherwise, $\pm$ is a Student-$t(4)$ 95\% half-width across the five resplit-level values; it measures variability across resplits of one corpus and is descriptive rather than a significance test. A \emph{win} is a resplit with a positive paired gap, so 5/5 reports consistency of direction within BOOM.

\paragraph{Zero-shot probe transfer} We apply the BOOM-trained coordination probe, with its BOOM weights and standardization and without refitting, to Toto activations on 11 FEV configurations and four LSTF datasets. The target is computed from each external context and its true next patch. The test asks only whether this fitted readout carries over unchanged; it does not measure forecast accuracy, and whether Toto's representations of external series encode coordination is a separate question for a probe refit on those series.

\section{Results}
\label{sec:results}

\subsection{Cadence and metric type are more recoverable from Toto than from input baselines}
\label{sec:structural}

Cadence and metric type are the two labels on which Toto's readout beats every baseline (Table~\ref{tab:structural-controls}). Against the strongest raw-window model, the paired gap is positive in all five resplits for both labels (Figure~\ref{fig:resplit-slopes}); it is also positive against the other three raw-window models in every resplit, and both labels beat random initialization and block permutation in 5/5. The cadence margin is the larger one, and it holds even though the FNO itself is well above the shuffled-label floor and the six-statistic probe; the metric-type margin is smaller but consistent across resplits.

\begin{table}[t]
\centering
\small
\caption{Common-support probes: held-out test macro-$F_1$ within one fixed stratum of the other two labels (means $\pm$ half-widths over five resplits; classes balanced by series). Test support is 11--19 (cadence), 17--26 (metric type), and 11--18 (domain) series per class. Wins are against random initialization and against the six-statistic probe.}
\label{tab:conditional}
\begin{tabularx}{\linewidth}{@{}l>{\raggedright\arraybackslash}Xcccc@{}}
\toprule
Target & Fixed stratum & Toto & Random init. & Six-stat. & Wins \\
\midrule
Cadence & Infrastructure/gauge & $0.697\pm0.060$ & $0.586\pm0.121$ & $0.485\pm0.137$ & 5/5; 5/5 \\
Metric type & Application Usage/Short & $0.502\pm0.053$ & $0.395\pm0.048$ & $0.371\pm0.034$ & 5/5; 5/5 \\
Domain & Short/gauge & $0.406\pm0.073$ & $0.350\pm0.091$ & $0.376\pm0.087$ & 4/5; 3/5 \\
\bottomrule
\end{tabularx}
\end{table}

\begin{table}[t]
\centering
\small
\setlength{\tabcolsep}{4pt}
\caption{MOMENT-base held-out test macro-$F_1$ on the same five resplits (means $\pm$ half-widths; views chosen on validation series). The last column is the paired gap to the six-statistic probe and the number of resplits in which it is positive.}
\label{tab:moment-structural}
\begin{tabular*}{\linewidth}{@{\extracolsep{\fill}}lcccc}
\toprule
Label & MOMENT & Random init. & Six-stat. & Gap to six-stat.\ (wins) \\
\midrule
Cadence & $0.632\pm0.113$ & $0.556\pm0.105$ & $0.401\pm0.099$ & $+0.231\pm0.044$ (5/5) \\
Metric type & $0.499\pm0.022$ & $0.452\pm0.029$ & $0.351\pm0.042$ & $+0.148\pm0.030$ (5/5) \\
Domain & $0.437\pm0.014$ & $0.409\pm0.020$ & $0.337\pm0.034$ & $+0.100\pm0.023$ (5/5) \\
Cardinality & $0.259\pm0.010$ & $0.257\pm0.018$ & $0.668\pm0.013$ & $-0.409\pm0.012$ (0/5) \\
\bottomrule
\end{tabular*}
\end{table}

Domain is nearly tied. Toto beats both backbone baselines, but its macro-$F_1$ is level with the GBDT's and its accuracy is lower ($0.534$ vs.\ $0.573$, 0/5 resplits), so a raw-window model recovers domain about as well.

Cardinality reverses the pattern. The CNN, FNO, and patch Transformer all reach macro-$F_1$ near $0.95$, far above Toto (App.~\ref{app:raw-controls}). These neural models see the variate count directly, through the zero-padded variate channels and the coverage channel. The two input baselines that see neither, the six-statistic probe and the GBDT, are much weaker than the neural models but still above Toto ($0.635$ and $0.663$ vs.\ $0.471$), so the reversal does not rest on the coverage channel alone.

\subsection{Common-support probes preserve two gaps in tested strata}
\label{sec:conditional}

Domain and metric type are moderately associated (series-level Cram\'er's $V=0.402\pm0.011$), whereas cadence is nearly independent of both ($0.055\pm0.012$ with domain and $0.050\pm0.018$ with metric type). Within fixed strata of the other two labels (Table~\ref{tab:conditional}), cadence and metric type keep positive gaps over the six-statistic probe and random initialization in every resplit, and their means stay well above the six-statistic probe. Domain's gaps are smaller and less consistent across resplits; it is also the label most associated with metric type, so composition and direct signal are hardest to separate for it. The strata are a targeted robustness check on small, balanced test support that varies by resplit. Rotated held-out-combination tests within BOOM are in App.~\ref{app:heldout-support}.

\subsection{MOMENT shows related taxonomy readouts on BOOM}
\label{sec:cross-model}

On the same resplits, MOMENT-base's cadence, metric-type, and domain readouts beat both its random initialization and the six-statistic probe in all five resplits (Table~\ref{tab:moment-structural}), although the margins over random initialization are modest and smallest for domain. Its cardinality readout matches random initialization and, like Toto's, trails the six-statistic probe. MOMENT's baselines are narrower than Toto's, since they exclude the raw-window models and block permutation, and its dynamic readouts are comparable to the six-statistic probe (App.~\ref{app:moment-details}). Cross-model recurrence on one benchmark does not rank architectures.

\begin{figure}[t]
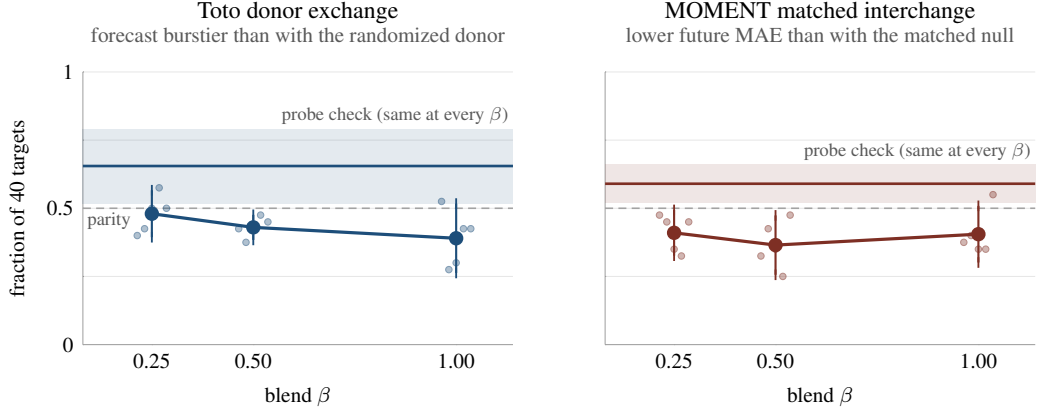

\centering
\input{figures/tikz/cr_style.tex}
\begin{tikzpicture}
\pgfplotsset{
  fig4 panel/.style={cr axis,
    scale only axis, width=2.24in, height=1.42in,
    xmin=0.08, xmax=1.14, ymin=0, ymax=1,
    xtick={0.25,0.5,1.0}, xticklabels={0.25,0.50,1.00},
    ytick={0,0.25,0.5,0.75,1}, yticklabels={0,,0.5,,1}, ymajorgrids=true,
    tick label style={font=\footnotesize}, xlabel={blend $\beta$}, xlabel style={font=\footnotesize},
    axis line style={crGrayDark, line width=0.4pt}, major tick length=0pt,
    title style={font=\small, align=center, text depth=0.4ex}, clip=false,
  },
  fig4 dots/.style={only marks, mark=*, mark size=1.2pt, color=\figFourColor,
    mark options={fill opacity=0.35, draw opacity=0.55}},
  fig4 mean/.style={color=\figFourColor, line width=1.2pt, mark=*, mark size=2.1pt,
    error bars/.cd, y dir=both, y explicit, error bar style={line width=0.8pt},
    error mark=-, error mark options={mark size=2pt, line width=0.7pt}},
}
\tikzset{
  fig4 band/.style={fill=\figFourColor, fill opacity=0.12},
  fig4 probe/.style={draw=\figFourColor, line width=1.0pt},
  fig4 lab/.style={font=\scriptsize, text=crMuted, inner sep=1.5pt},
}
\def\figFourColor{crToto}
\begin{axis}[fig4 panel, at={(0.42in,0in)}, anchor=south west,
  ylabel={fraction of 40 targets}, ylabel style={font=\footnotesize},
  title={Toto donor exchange\\[-1pt]{\footnotesize\color{crMuted}forecast burstier than with the randomized donor}}]
\draw[crGrayDark, line width=0.6pt, dash pattern=on 3pt off 2pt] (axis cs:0.08,0.5) -- (axis cs:1.14,0.5);
\node[fig4 lab, anchor=north west] at (axis cs:0.08,0.5) {parity};
\input{figures/tikz/gen/fig4_toto.tex}
\end{axis}
\def\figFourColor{crMoment}
\begin{axis}[fig4 panel, at={(3.14in,0in)}, anchor=south west, yticklabels={},
  title={MOMENT matched interchange\\[-1pt]{\footnotesize\color{crMuted}lower future MAE than with the matched null}}]
\draw[crGrayDark, line width=0.6pt, dash pattern=on 3pt off 2pt] (axis cs:0.08,0.5) -- (axis cs:1.14,0.5);
\input{figures/tikz/gen/fig4_moment.tex}
\end{axis}
\end{tikzpicture}
\caption{Probe check versus forecast endpoint across the residual blend $\beta$. The shaded band is the probe check (mean $\pm$ half-width), which is the same at every blend; points are resplits, and the line joins the forecast-endpoint means $\pm$ half-widths. The dashed line is parity.}
\label{fig:exchange}
\end{figure}

\subsection{Exchanges move the readout without consistently moving the forecast}
\label{sec:behavior}

The donor exchange asks whether moving the readout moves the forecast. In Toto, the probe check is above $0.5$ in all five resplits, so the exchange moves the future-burstiness readout as intended (Figure~\ref{fig:exchange}, left). The forecasts do not follow: the fraction of targets whose forecast is burstier under the high-burst than under the randomized donor is below one half on average at every blend and above it in at most one resplit (Table~\ref{tab:donor-exchange}). The secondary endpoint favors the high-burst donor at the point estimate, with median WAPE ratios below one, but its half-widths span one.

MOMENT's matched interchange patches its validation-selected layer-3, all-context view and draws its null donors from the nearest context-matched, non-high-burst windows of other series, with taxonomy penalties (App.~\ref{app:protocol}). It shows the same dissociation: the probe check is above parity in every resplit, while the fraction of targets with lower future MAE averages below parity at all three blends (Figure~\ref{fig:exchange}, right; Table~\ref{tab:moment-interchange}). Because MOMENT's null donors are matched to its high-burst donors on variate count, last-patch statistics, and taxonomy, its contrast holds these donor properties approximately fixed. Toto's randomized donor is not matched, so Toto's contrast may also reflect donor differences other than burstiness.

Toto's future-burstiness probe reaches $R^2=0.106\pm0.100$ against $0.111\pm0.066$ for the six-statistic probe (above it in 3/5 resplits), so this label is about as recoverable from simple input statistics as from Toto's residuals. Each diagnostic also contrasts two exchanges with each other rather than with an intact forecast, so it measures relative forecast response, not a causal direction the model uses naturally.

\subsection{Zero-shot coordination-probe transfer is negative}
\label{sec:transfer}

Applied without refitting, the BOOM-trained coordination probe has negative $R^2$ on both external suites, so it does worse than predicting the mean, by a much larger margin on LSTF than on FEV (Table~\ref{tab:external-transfer}). We report only coordination because several other external dynamic labels are nearly constant, which makes their $R^2$ unstable. This is a failure of zero-shot probe transfer: the fitted weights and BOOM standardization do not carry over to these series. It is not evidence that Toto's representations of external series lack coordination information, which only a probe refit on those series could test. The positive evidence in this paper is scoped to BOOM.

\section{Discussion}
\label{sec:discussion}

\paragraph{A label-by-label profile}
The results describe what is linearly recoverable from Toto's residuals, label by label, rather than giving a single verdict on what the model ``knows''. Cadence has the largest advantage over its strongest raw-window model, metric type a smaller but consistent one, domain is nearly tied, and cardinality is far better recovered from the raw window. These statements concern the evaluated representation, probe family, and baselines, not the total information the model contains: a nonlinear probe, another view, or a stronger raw-window model could shift any of them.

\paragraph{Why cardinality differs}
In every resplit, Toto's validation-selected cardinality view is the layer-11, all-context, series-mean representation, which averages over variates and carries no explicit count, whereas the neural raw-window models see the count directly. Architecture, input format, and readout capacity also differ, so mean pooling is one candidate cause among several. The comparison shows that a label's accessibility can change sharply with the representation and input given to the evaluator.

\paragraph{What cadence may reflect}
One hypothesis is that BOOM's Short/Medium metadata tracks temporal regularities visible across Toto's patch context. Since cadence is a metadata label rather than a spectral estimate, this mechanism is left to controlled tests.

\paragraph{Readout is not use}
In both models, exchanges move the probe readout while forecasts do not respond consistently. Related TSFM work has used trend and periodicity concepts to steer forecasts \citep{wilinski2025representations}, and synthetic-factor studies probe temporal representations under controlled conditions \citep{pandey2025internalsemantics}. Our exchange differs from concept-direction steering in both the concept (future burstiness) and the intervention (cross-series donor residuals), so its dissociation characterizes this exchange and endpoint and is compatible with prior steering results. The next Toto test would match donors on context and taxonomy as MOMENT's interchange does, preserve the target context while changing the intended factor, include sham exchanges, and report several forecast endpoints.

\paragraph{Limits and next steps}
Five resplits of one corpus measure split variability, not uncertainty over independent observability data, and unknown pretraining overlap leaves the source of the signal unattributed. The cadence label covers Short and Medium series, and the common-support probes omit the raw-window models. A next study should repeat the paired design across independent telemetry corpora and TSFMs, with comparable raw-window models and label definitions, broader cadence classes, raw-window models refit within strata, and independently sourced series; controlled tests that vary context and patch scales would help explain the cadence result.

\paragraph{Recommendations for practice}
A probe score on a forecasting model is interpretable only relative to the strongest baseline it beats. For each label, we recommend reporting the strongest input baseline, whether the readout also beats the same architecture without its trained configuration, and, separately, whether moving the readout moves the forecast; a claim should go no further than the comparisons it passes. Here, cadence and metric type pass the first two. The forecast test, run on future burstiness because its next-patch definition fixes an intended forecast direction, did not show a consistent forecast change.

\paragraph{Broader impacts}
The audit helps researchers separate recoverable information from evidence of use: a probe score treated as a steering target could produce unintended forecast changes, and the donor-exchange readout is not a validated forecast-control knob. Reporting representation and forecast outcomes side by side keeps that boundary visible.

\section{Conclusion}
\label{sec:conclusion}

Across five series-disjoint BOOM resplits, short-vs-medium cadence and metric type are more linearly recoverable from Toto's frozen residuals than from every tested raw-window model, and their gaps over the six-statistic probe and random initialization persist in the tested strata. Domain is nearly tied with its strongest raw-window model, and cardinality is far better recovered from the raw window. MOMENT-base shows cadence, metric-type, and domain readouts above random initialization and the six-statistic probe on the same benchmark. Exchanging residuals moves the targeted readout without a consistent forecast change, and the BOOM-trained coordination probe has negative zero-shot $R^2$ on the tested external benchmarks. Reporting each readout against its strongest baseline, alongside forecast and transfer tests, keeps the line between what a model makes recoverable and what it uses explicit.

Code and audited results are publicly available at \url{https://github.com/DhyeyMavani2003/toto-interp/tree/neurips-2026-v1}.

\begin{ack}
\textbf{Funding.} This work received no external funding or third-party support.

\textbf{Competing interests.} The authors declare no competing interests.
\end{ack}

\bibliographystyle{plainnat}
\bibliography{references}

\newpage
\appendix

\section{Protocol Details}
\label{app:protocol}

\paragraph{Data and splits}
Each resplit (seeds 42--46) partitions the $2{,}807$ BOOM series 70/15/15 by seeded shuffle with training capped at the first 500 shuffled training series, giving 500/421/422 train/validation/test series per resplit. Up to four windows per series are sampled at integer linspace starts over valid starts, excluding the latest start. Context is 1024 observations with a 64-observation patch; the target is the next patch. MOMENT-suite window counts per resplit span 1224--1296 train, 984--1077 validation, and 1044--1074 test windows; the Toto donor-exchange pool holds 1032--1068 test windows.

\paragraph{Views and probes}
Candidate views are 13 layers (patch embedding plus blocks 0--11) by 3 token positions (all-context, final-context, first-decode) with series-mean pooling: 39 views for pretrained and random-initialization probes; the block-permutation grid additionally admits per-variate pooling (78 views). Categorical probes are logistic regression (balanced class weights, L2, $C=1$, lbfgs, 4000 iterations); continuous probes are ridge ($\alpha=1$). Features use train-split standardization (standard deviation floored at $10^{-6}$; NaN to 0, infinities to $\pm 10^{3}$, clipped). Only the view is selected, by validation accuracy (macro-$F_1$ for cardinality); there is no regularization search.

\paragraph{Baselines}
The six-statistic probe uses the mean, standard deviation, zero fraction, mean absolute value, and maximum absolute value over the last context patch, plus the standard deviation of per-variate means. The shuffled-label probe shuffles train-split targets with the resplit seed, keeps the same features and estimator, and is evaluated on unshuffled validation and test. Random initialization resets the full backbone with the resplit seed and is evaluated frozen. Block permutation reorders the 12 blocks by seeded randperm with identity rejection; runtime orders are 42: [6, 8, 1, 7, 0, 2, 10, 11, 4, 3, 5, 9]; 43: [4, 6, 7, 2, 9, 8, 1, 5, 0, 10, 11, 3]; 44: [4, 5, 9, 3, 7, 1, 8, 6, 11, 0, 10, 2]; 45: [11, 6, 3, 0, 4, 5, 7, 10, 2, 1, 8, 9]; 46: [9, 5, 4, 0, 6, 8, 2, 7, 3, 1, 11, 10]. Raw-window models (Table~\ref{tab:raw-window-models}) train on padded contexts times a variate mask plus a coverage channel (neural models) or 15 features (GBDT: 6 raw, 6 spectral, 3 autocorrelation). Neural models use AdamW (lr $10^{-3}$, weight decay $10^{-4}$), 20 epochs, batch 16, weighted cross-entropy, and the best-validation-epoch checkpoint (macro-$F_1$ categorical, $R^2$ continuous). GBDT is HistGradientBoosting at a fixed 300 iterations with no validation selection.

\begin{table}[ht]
\centering
\small
\setlength{\tabcolsep}{3pt}
\caption{Raw-window baseline models. Neural models see the masked full context plus a coverage channel; GBDT sees 15 engineered features.}
\label{tab:raw-window-models}
\footnotesize
\begin{tabular}{lcccc}
\toprule
Model & Parameters & Input & Training & Checkpoint selection \\
\midrule
CNN & 39.5k & masked ctx + coverage & AdamW, 20 ep, batch 16 & best val epoch \\
FNO & 105.9k & masked ctx + coverage & AdamW, 20 ep, batch 16 & best val epoch \\
Patch Transformer & 246.7k & masked ctx + coverage & AdamW, 20 ep, batch 16 & best val epoch \\
GBDT & -- & 15 eng.\ features & HistGradBoost, 300 iter & none (fixed budget) \\
\bottomrule
\end{tabular}
\end{table}

\paragraph{Label definitions}
Dynamic labels use all target variates of the series (which for BOOM windows is every variate; there are no held-out covariates); the last context patch is the final $P$ observations ($P=64$ Toto, $P=8$ MOMENT); the robust scale is $1.4826$ times the median absolute deviation, floored at $10^{-6}$. Taxonomy: cardinality bins are 1, 2--8, 9--32, and above 32; metric type and domain keep singleton primary entries and map other/multilabel rows to auxiliary, which are dropped from fits; cadence is the raw BOOM frequency string (the Long bucket yields no 1024-context window).

\paragraph{Donor exchanges}
Toto donor-exchange targets are test windows in the lowest future-burstiness quartile; high-burst donors are top-quartile windows from other series; randomized donors come from all eligible test windows of other series with no covariate or taxonomy matching. Each donor window is used at most once per arm; the same 40 triples are reused across blends. Forecasts use 16 samples over the 64-observation next patch with CPU/CUDA RNG state shared across arms. Forecast burstiness is the maximum over variates and steps of \mbox{$|\mathrm{forecast\ median}-\mathrm{context\ median}|/\mathrm{robust\ scale}$}; WAPE is $\sum_{i,t}|y_{i,t}-\hat{y}_{i,t}|/\sum_{i,t}|y_{i,t}|$ over target variates $i$ and forecast steps $t$ (denominator floored at $10^{-6}$). For BOOM windows the target variates are all variates of the series, with no held-out covariates. MOMENT's matched interchange instead patches its validation-selected layer-3 all-context view: real donors have future burstiness above the 0.75 quantile with the target's variate count; matched null donors are below it, drawn from the $K=5$ nearest context-matched candidates with taxonomy penalties. Its forecast endpoint is future MAE from the official short-forecast head.

\paragraph{Zero-shot probe transfer}
The BOOM-trained probe (weights plus BOOM-train standardization) is applied without refitting to external activations at the same view: one validation-selected probe per continuous label, context 1024, up to 100 series by 4 windows per dataset across 11 FEV configurations (\texttt{entsoe\_15T}, \texttt{entsoe\_30T}, \texttt{entsoe\_1H}, \texttt{epf\_be}, \texttt{epf\_de}, \texttt{epf\_fr}, \texttt{epf\_np}, \texttt{epf\_pjm}, \texttt{rohlik\_sales\_1D}, \texttt{solar\_with\_weather\_15T}, \texttt{solar\_with\_weather\_1H}) and 4 LSTF datasets (ETTh1, ETTh2, weather, electricity). The FEV configurations come from ENTSO-E Transparency Platform load data via Open Power System Data (\texttt{entsoe}) \citep{opsd2020timeseries}, the open-access day-ahead electricity-price benchmark (\texttt{epf}) \citep{lago2021epf}, the Rohlik Sales Forecasting Challenge data (\texttt{rohlik\_sales\_1D}) \citep{kecera2024rohlik}, and a Kaggle solar and weather dataset (\texttt{solar\_with\_weather}) \citep{saman2023solar}. ETTh1/ETTh2 come from ETDataset \citep{zhou2021informer}, electricity from the UCI ElectricityLoadDiagrams20112014 data \citep{trindade2015electricity}, and weather from the Max Planck Institute for Biogeochemistry weather records in Jena \citep{mpibgc2020weather}. Within each resplit, transfer averages dataset-level $R^2$ with equal weights.

\paragraph{Compute}
The main five-resplit suite ran on a CPU cluster (16 CPUs; about 1:07--1:25 elapsed, 18.7--21.3 GB peak RSS). GPU stages on NVIDIA A10 nodes took 21--37 min (28.6--29.1 GB) for MOMENT activations, 3--4 min (6.7--9.0 GB) for the Toto donor exchange, and about 1--2 min (1.7--3.0 GB) for the MOMENT matched interchange.

\section{Within-BOOM Held-Out-Combination Stress Tests}
\label{app:heldout-support}

We also evaluated rotated held-out-combination tests: in each resplit, a support-valid domain or taxonomy combination was withheld from both training and validation, then the matching test subset was evaluated. Because different values were withheld across resplits, the brackets in Table~\ref{tab:heldout-support} are coverage ranges across held-out values, not confidence intervals for a shared population. These tests characterize within-BOOM coverage under the tested support.

\begin{table}[ht]
\centering
\small
\setlength{\tabcolsep}{4pt}
\caption{Rotated within-BOOM held-out-combination tests. Each bracket spans scores from distinct held-out values across the five resplits; it is not an uncertainty interval.}
\label{tab:heldout-support}
\begin{tabular}{@{}l>{\raggedright\arraybackslash}p{2.8cm}@{\hspace{14pt}}>{\centering\arraybackslash}p{3.5cm}@{\hspace{14pt}}>{\centering\arraybackslash}p{3.5cm}@{}}
\toprule
Target & \shortstack[l]{Held-out\\support} & \shortstack{Toto macro-$F_1$\\range} & \shortstack{Random init.\\range} \\
\midrule
Cadence & Domain $\times$ metric type & $0.865\ [0.627,1.000]$ & $0.541\ [0.250,1.000]$ \\
Cadence & Domain & $0.654\ [0.513,0.766]$ & $0.511\ [0.420,0.575]$ \\
Metric type & Domain $\times$ cadence & $0.258\ [0.130,0.472]$ & $0.179\ [0.135,0.224]$ \\
Metric type & Domain & $0.281\ [0.188,0.505]$ & $0.220\ [0.161,0.266]$ \\
\bottomrule
\end{tabular}
\end{table}

\section{Additional Diagnostic Tables}
\label{app:moment-details}

Table~\ref{tab:donor-exchange} gives the Toto donor-exchange endpoints summarized in \S\ref{sec:behavior} and Figure~\ref{fig:exchange}; Table~\ref{tab:external-transfer} gives the zero-shot probe transfer results of \S\ref{sec:transfer}; Tables~\ref{tab:moment-dynamic} and~\ref{tab:moment-interchange} report MOMENT's dynamic readouts and matched interchange.

\begin{table}[!ht]
\centering
\small
\caption{Toto donor exchange: fraction of the 40 targets per resplit whose forecast is burstier under the high-burst than under the randomized donor, and the median WAPE ratio (high-burst/randomized). Means $\pm$ half-widths across resplits; the same triples are used at every blend. Donors come from other series and are not matched on covariates or taxonomy. Probe check: $0.655\pm0.136$ at every blend.}
\label{tab:donor-exchange}
\begin{tabular*}{\linewidth}{@{\extracolsep{\fill}}cccc}
\toprule
Blend $\beta$ & Forecast burstier & Resplits $>0.5$ & Median WAPE ratio \\
\midrule
0.25 & $0.480\pm0.086$ & 1/5 & $0.969\pm0.062$ \\
0.50 & $0.430\pm0.046$ & 0/5 & $0.952\pm0.108$ \\
1.00 & $0.390\pm0.127$ & 1/5 & $0.903\pm0.155$ \\
\bottomrule
\end{tabular*}
\end{table}

\begin{table}[!ht]
\centering
\small
\caption{Zero-shot coordination-probe transfer (BOOM-trained probe, no refitting). $R^2$ is macro-averaged over the datasets of each set within a resplit, relative to a constant-mean reference; means $\pm$ half-widths across resplits.}
\label{tab:external-transfer}
\begin{tabular}{lcr@{\,$\pm$\,}l}
\toprule
Benchmark set & Entries & \multicolumn{2}{c}{Coordination $R^2$} \\
\midrule
FEV & 11 & $-0.938$ & $0.654$ \\
LSTF & 4 & $-17.668$ & $10.896$ \\
\bottomrule
\end{tabular}
\end{table}

\begin{table}[!ht]
\centering
\small
\setlength{\tabcolsep}{4pt}
\caption{MOMENT dynamic readouts on held-out BOOM series. Views are selected on validation $R^2$ and evaluated once on test series. The near-constant sparsity targets are omitted; their six-statistic $R^2$ values are unstable. Means $\pm$ half-widths (§\ref{sec:method}).}
\label{tab:moment-dynamic}
\begin{tabular}{lr@{\,$\pm$\,}lr@{\,$\pm$\,}lr@{\,$\pm$\,}lc}
\toprule
Dynamic label & \multicolumn{2}{c}{MOMENT $R^2$} & \multicolumn{2}{c}{Random init.} & \multicolumn{2}{c}{Six-stat.} & MOMENT above six-stat. \\
\midrule
Coordination & $0.092$ & $0.047$ & $0.083$ & $0.041$ & $0.071$ & $0.041$ & 3/5 \\
Current burstiness & $0.024$ & $0.025$ & $0.007$ & $0.018$ & $0.057$ & $0.620$ & 2/5 \\
Future burstiness & $0.015$ & $0.027$ & $0.002$ & $0.021$ & $-0.005$ & $0.082$ & 2/5 \\
Shift risk & $-0.006$ & $0.067$ & $-0.018$ & $0.062$ & $0.021$ & $0.076$ & 1/5 \\
\bottomrule
\end{tabular}
\end{table}

\needspace{5\baselineskip}
MOMENT's dynamic readouts are comparable to the six-statistic probe (Table~\ref{tab:moment-dynamic}). Coordination is above it in 3/5 resplits and future burstiness in 2/5; the six-statistic probe is highly variable. The cross-model readout claim therefore covers cadence, metric type, and domain. For the matched interchange (Table~\ref{tab:moment-interchange}), the probe check is blend-invariant and reported in the caption; only the forecast endpoint varies by blend.

\begin{table}[!ht]
\centering
\small
\setlength{\tabcolsep}{4pt}
\caption{MOMENT matched interchange. Each resplit evaluates 40 held-out target/real/matched-null triples with the same triples reused at each blend. Probe check (blend-invariant): $0.590\pm0.071$, above 0.5 in 5/5 resplits. The lower future-MAE fraction is the forecast endpoint. Means $\pm$ half-widths (§\ref{sec:method}).}
\label{tab:moment-interchange}
\begin{tabular}{ccc}
\toprule
Blend $\beta$ & Lower future-MAE fraction & Resplits $>0.5$ \\
\midrule
0.25 & $0.410\pm0.084$ & 0/5 \\
0.50 & $0.365\pm0.109$ & 0/5 \\
1.00 & $0.405\pm0.104$ & 1/5 \\
\bottomrule
\end{tabular}
\end{table}

\needspace{0.33\textheight}
\section{All Raw-Window and Backbone Baselines}
\label{app:raw-controls}

Table~\ref{tab:structural-controls} reports the strongest raw-window model for each label; Table~\ref{tab:raw-control-matrix} lists every neural and tree raw-window model alongside the two backbone baselines, random initialization and block permutation.

\begin{table}[!ht]
\centering
\small
\setlength{\tabcolsep}{3pt}
\caption{Held-out test macro-$F_1$ of Toto and all raw-window and backbone baselines over five resplits. Neural raw-window checkpoints are selected on validation data. Means $\pm$ half-widths (§\ref{sec:method}).}
\label{tab:raw-control-matrix}
\begin{tabular*}{\linewidth}{@{\extracolsep{\fill}}lccc}
\toprule
Target & Toto & Random init. & Block perm. \\
\midrule
Cadence & $0.766\pm0.024$ & $0.547\pm0.040$ & $0.567\pm0.020$ \\
Metric type & $0.545\pm0.037$ & $0.398\pm0.035$ & $0.420\pm0.016$ \\
Domain & $0.484\pm0.011$ & $0.362\pm0.017$ & $0.394\pm0.011$ \\
Cardinality & $0.471\pm0.029$ & $0.250\pm0.025$ & $0.428\pm0.127$ \\
\end{tabular*}

\vspace{3pt}
\begin{tabular*}{\linewidth}{@{\extracolsep{\fill}}lcccc}
\toprule
Target & CNN & FNO & \shortstack{Patch\\Transformer} & GBDT \\
\midrule
Cadence & $0.545\pm0.019$ & $0.633\pm0.047$ & $0.553\pm0.027$ & $0.580\pm0.042$ \\
Metric type & $0.431\pm0.047$ & $0.366\pm0.061$ & $0.398\pm0.018$ & $0.498\pm0.007$ \\
Domain & $0.432\pm0.025$ & $0.381\pm0.069$ & $0.437\pm0.029$ & $0.482\pm0.040$ \\
Cardinality & $0.959\pm0.010$ & $0.961\pm0.016$ & $0.948\pm0.023$ & $0.663\pm0.038$ \\
\bottomrule
\end{tabular*}
\end{table}

\FloatBarrier
\section*{NeurIPS Paper Checklist}

\begin{enumerate}[itemsep=0pt,topsep=0pt,parsep=0pt]
\needspace{6\baselineskip}
\item {\bf Claims}
    \item[] Question: Do the main claims made in the abstract and introduction accurately reflect the paper's contributions and scope?
    \item[] Answer: \answerYes{}
    \item[] Justification: The abstract and introduction state the claim as linear recoverability from Toto's frozen residuals, identify cadence and metric type as the labels more recoverable from those residuals than from the tested input and backbone baselines on BOOM, and scope the domain, cardinality, forecast, and zero-shot probe transfer claims by their baselines (\S\ref{sec:results}).

\needspace{6\baselineskip}
\item {\bf Limitations}
    \item[] Question: Does the paper discuss the limitations of the work performed by the authors?
    \item[] Answer: \answerYes{}
    \item[] Justification: The paper scopes its claims to five resplits of one BOOM corpus with unknown pretraining overlap, states the narrower baselines of the common-support and MOMENT comparisons, and reports readout--forecast dissociation and negative zero-shot probe transfer (\S\ref{sec:setup}, \S\ref{sec:method}, \S\ref{sec:conditional}, \S\ref{sec:cross-model}, \S\ref{sec:behavior}, \S\ref{sec:transfer}, \S\ref{sec:discussion}).

\needspace{6\baselineskip}
\item {\bf Theory assumptions and proofs}
    \item[] Question: For each theoretical result, does the paper provide the full set of assumptions and a complete (and correct) proof?
    \item[] Answer: \answerNA{}
    \item[] Justification: The paper is empirical and presents no formal theorems.

\needspace{6\baselineskip}
\item {\bf Experimental result reproducibility}
    \item[] Question: Does the paper fully disclose all the information needed to reproduce the main experimental results of the paper to the extent that it affects the main claims and/or conclusions of the paper (regardless of whether the code and data are provided or not)?
    \item[] Answer: \answerYes{}
    \item[] Justification: The paper and App.~\ref{app:protocol} specify the model checkpoints, data sources, resplits and seeds, candidate views, probe settings, baselines, donor exchanges, and zero-shot probe transfer protocol needed to reproduce the main results.

\needspace{6\baselineskip}
\item {\bf Open access to data and code}
    \item[] Question: Does the paper provide open access to the data and code, with sufficient instructions to faithfully reproduce the main experimental results, as described in supplemental material?
    \item[] Answer: \answerYes{}
    \item[] Justification: The code, audited result files, a CPU verifier for the reported tables and figures, and setup and GPU rerun instructions are publicly available at \url{https://github.com/DhyeyMavani2003/toto-interp/tree/neurips-2026-v1}. The repository contains no dataset files or model weights; its scripts download BOOM, FEV, Toto, and MOMENT from Hugging Face and LSTF from the Time-Series-Library links.

\needspace{6\baselineskip}
\item {\bf Experimental setting/details}
    \item[] Question: Does the paper specify all the training and test details (e.g., data splits, hyperparameters, how they were chosen, type of optimizer) necessary to understand the results?
    \item[] Answer: \answerYes{}
    \item[] Justification: The paper specifies the checkpoints, five seeded series-disjoint resplits, labels, activation views, probes, baselines, and donor exchanges (\S\ref{sec:setup}, \S\ref{sec:method}, \S\ref{sec:results}, App.~\ref{app:protocol}).

\needspace{6\baselineskip}
\item {\bf Experiment statistical significance}
    \item[] Question: Does the paper report error bars suitably and correctly defined or other appropriate information about the statistical significance of the experiments?
    \item[] Answer: \answerYes{}
    \item[] Justification: The Student-$t(4)$ half-widths describe variation across five resplits of one BOOM corpus, not significance tests or independent-corpus intervals; donor triples are reused across blends (\S\ref{sec:method}, \S\ref{sec:results}, App.~\ref{app:protocol}).

\needspace{6\baselineskip}
\item {\bf Experiments compute resources}
    \item[] Question: For each experiment, does the paper provide sufficient information on the computer resources (type of compute workers, memory, time of execution) needed to reproduce the experiments?
    \item[] Answer: \answerYes{}
    \item[] Justification: The Compute paragraph of App.~\ref{app:protocol} reports the CPU and GPU allocations, elapsed times, and peak memory for the main five-resplit stages.

\needspace{6\baselineskip}
\item {\bf Code of ethics}
    \item[] Question: Does the research conducted in the paper conform, in every respect, with the NeurIPS Code of Ethics \url{https://neurips.cc/public/EthicsGuidelines}?
    \item[] Answer: \answerYes{}
    \item[] Justification: The work uses existing model checkpoints and benchmark data, involves no human-subject research, and makes no deployment-safety or fairness guarantee.

\needspace{6\baselineskip}
\item {\bf Broader impacts}
    \item[] Question: Does the paper discuss both potential positive societal impacts and negative societal impacts of the work performed?
    \item[] Answer: \answerYes{}
    \item[] Justification: The discussion states that the audit helps separate recoverable information from evidence of use, that a probe score treated as a steering target could produce unintended forecast changes, and that the donor-exchange readout is not a validated forecast-control knob (\S\ref{sec:discussion}).

\needspace{6\baselineskip}
\item {\bf Safeguards}
    \item[] Question: Does the paper describe safeguards that have been put in place for responsible release of data or models that have a high risk for misuse (e.g., pre-trained language models, image generators, or scraped datasets)?
    \item[] Answer: \answerNA{}
    \item[] Justification: The work introduces no new forecasting model or dataset; it evaluates published models, describes an audit protocol, and does not present its exchanges as deployment controls.

\needspace{6\baselineskip}
\item {\bf Licenses for existing assets}
    \item[] Question: Are the creators or original owners of assets (e.g., code, data, models), used in the paper, properly credited and are the license and terms of use explicitly mentioned and properly respected?
    \item[] Answer: \answerYes{}
    \item[] Justification: All assets and their original sources are cited (\S\ref{sec:setup}, App.~\ref{app:protocol}). Licenses: Toto-Open-Base-1.0 and BOOM (Apache-2.0), MOMENT-1-base (MIT), ETDataset (CC BY-ND 4.0), and the UCI electricity and MPI-BGC Jena weather data (CC BY 4.0). The FEV configurations are used under FEV's research-use terms and credited to their original sources. No dataset or model weights are redistributed.

\needspace{6\baselineskip}
\item {\bf New assets}
    \item[] Question: Are new assets introduced in the paper well documented and is the documentation provided alongside the assets?
    \item[] Answer: \answerYes{}
    \item[] Justification: We release analysis code and aggregate experimental results. The repository linked in item 5 documents the protocol, dependencies, execution steps, compute needs, limitations, and license terms. No new dataset or model weights are released.

\needspace{6\baselineskip}
\item {\bf Crowdsourcing and research with human subjects}
    \item[] Question: For crowdsourcing experiments and research with human subjects, does the paper include the full text of instructions given to participants and screenshots, if applicable, as well as details about compensation (if any)? 
    \item[] Answer: \answerNA{}
    \item[] Justification: The work includes no crowdsourcing or human-subject research.

\needspace{6\baselineskip}
\item {\bf Institutional review board (IRB) approvals or equivalent for research with human subjects}
    \item[] Question: Does the paper describe potential risks incurred by study participants, whether such risks were disclosed to the subjects, and whether Institutional Review Board (IRB) approvals (or an equivalent approval/review based on the requirements of your country or institution) were obtained?
    \item[] Answer: \answerNA{}
    \item[] Justification: The work includes no human-subject research.

\needspace{6\baselineskip}
\item {\bf Declaration of LLM usage}
    \item[] Question: Does the paper describe the usage of LLMs if it is an important, original, or non-standard component of the core methods in this research? Note that if the LLM is used only for writing, editing, or formatting purposes and does \emph{not} impact the core methodology, scientific rigor, or originality of the research, declaration is not required.
    \item[] Answer: \answerNA{}
    \item[] Justification: LLMs are not part of the research method, data processing, or experiments described in the paper.

\end{enumerate}

\end{document}